\documentclass[sigconf]{acmart}

\makeatletter                   
\def\mdseries@tt{m}             
\makeatother                    

\usepackage{underscore}           
\usepackage{amsmath}
\usepackage{latexsym}
\usepackage{subcaption}
\usepackage{algorithm}
\usepackage{algpseudocode}
\usepackage{multirow}
\usepackage{multicol}
\usepackage{booktabs}
\usepackage{listings}
\usepackage{graphicx}
\usepackage{dblfloatfix}
\usepackage{hyperref}           
\hypersetup{
    colorlinks=true,
    linkcolor=blue,
    filecolor=red,      
    urlcolor=magenta,
    breaklinks=true,            
}
\usepackage{breakurl}           
\usepackage{booktabs}

\setcopyright{none}

\renewcommand\footnotetextcopyrightpermission[1]{} 
\newif\ifsubmit
\submitfalse
\ifsubmit
\newcommand{\sknote}[1]{}

\else
\newcommand{\sknote}[1]{\textcolor{blue}{\textbf{Sidd: #1}}}

\fi

\algnewcommand{\IfThen}[2]{
  \State \algorithmicif\ #1\ \algorithmicthen\ #2}

\makeatletter
\renewcommand\fs@ruled{%
  \def\@fs@cfont{\rmfamily}%
  \let\@fs@capt\floatc@plain%
  \def\@fs@pre{\hrule height.8pt depth0pt \kern2pt}
  \def\@fs@post{}
  \def\@fs@mid{\kern2pt\hrule\kern2pt}
  \let\@fs@iftopcapt\iffalse}
\makeatother

\begin{document}
\sloppy
\title{Improving Grey-Box Fuzzing by Modeling Program Behavior}

\author{Siddharth Karamcheti}
\authornote{Work completed while an intern at Bloomberg}
\affiliation{%
	\institution{Bloomberg}
    \department{CTO Data Science}
    \city{New York}
    \state{NY}
    \country{USA}
}
\email{sidd.karamcheti@gmail.com}
\email{}

\author{Gideon Mann}
\affiliation{%
	\institution{Bloomberg}
    \department{CTO Data Science}
    \city{New York}
    \state{NY}
    \country{USA}
}
\email{gmann16@bloomberg.net}

\author{David Rosenberg}
\affiliation{%
	\institution{Bloomberg}
    \department{CTO Data Science}
    \city{New York}
    \state{NY}
    \country{USA}
}
\email{drosenberg44@bloomberg.net}

\begin{abstract}
Grey-box fuzzers such as American Fuzzy Lop (AFL) are popular tools for finding bugs and potential vulnerabilities in programs. While these fuzzers have been able to find vulnerabilities in many widely used programs, they are not efficient; of the millions of inputs executed by AFL in a typical fuzzing run, only a handful discover unseen behavior or trigger a crash. The remaining inputs are redundant, exhibiting behavior that has already been observed. Here, we present an approach to increase the efficiency of fuzzers like AFL by applying machine learning to directly model how programs behave. We learn a forward prediction model that maps program inputs to execution traces, training on the thousands of inputs collected during standard fuzzing. This learned model guides exploration  by focusing on fuzzing inputs on which our model is the most uncertain (measured via the entropy of the predicted execution trace distribution). By focusing on executing inputs our learned model is unsure about, and ignoring any input whose behavior our model is certain about, we show that we can significantly limit wasteful execution. Through testing our approach on a set of binaries released as part of the DARPA Cyber Grand Challenge, we show that our approach is able to find a set of inputs that result in more code coverage and discovered crashes than baseline fuzzers with significantly fewer executions. 

\end{abstract}

\ccsdesc[500]{Software and its engineering~Software testing and debugging}

\ccsdesc[100]{Security and privacy~Software security engineering}

\keywords{program modeling; binary fuzzing; coverage-based fuzzing}

\maketitle

\section{Introduction}

The goal of fuzz-testing, or fuzzing, is to discover a set of test inputs that maximize code coverage in a given program, with the hope that doing so allows one to find bugs, crashes, or other potential vulnerabilities. While there are many tools for fuzzing, grey-box mutational fuzzers such as American Fuzzy Lop (AFL) are among the most successful. These fuzzers work by maintaining a queue of interesting program inputs, or ``parents'', that cover different parts of the program, and mutating them iteratively, with a set of stochastic mutation functions (e.g. flip bits, delete bits, insert random bits, etc.) to generate new ``children'' inputs. These children are then fed to a version of the program that has been lightly instrumented to trace the execution for a given input. If the input takes a path through the program that has not been observed before, it is added to the queue. Otherwise, it is discarded. Unfortunately, discarding inputs comes at a cost; each execution takes time, ranging from a couple of nanoseconds, to longer than a second, depending on the program. Within a typical fuzzing run, on the order of billions of inputs are generated, with only a handful actually covering unseen code paths, leading to hundreds of minutes of unnecessary execution time. In this work, we propose a method to cut down on these redundant executions by using machine learning to model program behavior.



Specifically, we posit that to fuzz successfully, it is important to be able to correlate program inputs with the resulting execution paths. By using machine learning to predict the execution path from a given input, we introduce an approach that is complementary to grey-box fuzzing, allowing us to filter useless inputs prior to execution. The intuition behind our filtering approach is simple: we focus on executing the generated inputs on which our learned reasoning model expresses low confidence (if we cannot reason about what the input will do, chances are it is likely to do something different than what we have seen before). By focusing on \textbf{modeling program behavior} we show that we can significantly improve program coverage with a smaller number of program executions.

We build our approach on top of AFL, one of the preeminent grey-box fuzzers. We show through a series of experiments on the DARPA Cyber Grand Challenge binaries that our approach offers significant improvements to fuzzing efficiency, obtaining a higher rate of code coverage than many strong baselines, including the best performing version of AFL.

\begin{figure*}
    \centering
    \includegraphics[scale=.75]{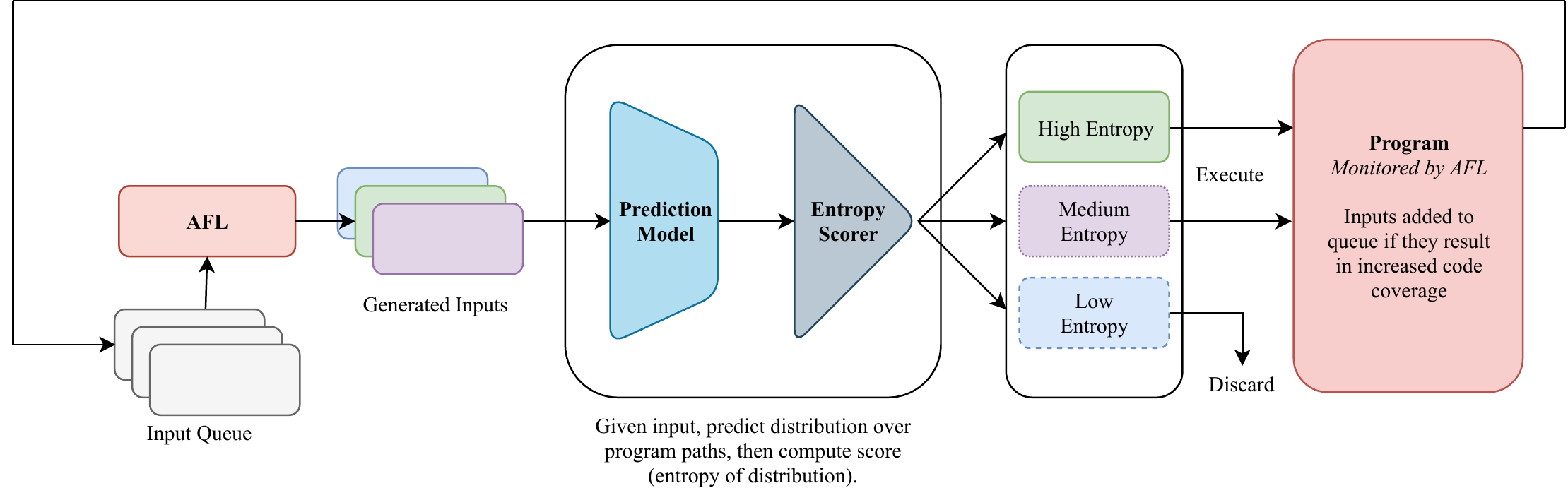}
    \caption{Full system pipeline. We use AFL to generate a number of test inputs, then rank these test inputs with our prediction model and entropy scorer. We then execute the inputs with the highest scores, and discard the rest.}
    \label{fig:pipeline}
\end{figure*}

\section{Related Work}

While there are many classes of approaches for finding bugs in programs, we focus on two key types: white-box approaches \cite{Ganesh2009TaintbasedDW}, including symbolic execution \cite{Li2013SteeringSE,Ma2011DirectedSE,Baldoni2018ASO,Pasareanu2011SymbolicEW,Wang2017AngrT,David2016BINSECSEAD, Cadar2008KLEEUA, Godefroid2012SAGEWF}, and grey-box approaches \cite{Rawat2017VUzzerAE,Stephens2016DrillerAF,AFL,Lemieux2017FairFuzzTR,Bhme2016CoveragebasedGF,Li2017SteelixPB}. These names come from the amount of transparency into the underlying program required; white-box approaches require a great deal of transparency (often access to the program source, or the ability to lift a binary up to an intermediate representation), while grey-box approaches require very little (often it's just enough to instrument the program at runtime, to collect small pieces of information, such as whenever the code enters a new basic block). There also exist black-box approaches \cite{Woo2013SchedulingBM,Gascon2015PulsarSB,Radamsa,ZZUF} that randomly generate inputs very quickly, and execute programs to discover crashes. While useful for finding shallow bugs, they often fail to penetrate deeply into programs.

The core of white-box approaches is the ability to leverage full transparency into the program under test, in order to explicitly reason about the nature of control flow. Symbolic execution tools like KLEE \cite{Cadar2008KLEEUA} and SAGE \cite{Godefroid2012SAGEWF} lift programs up to an intermediate representation, where the entirety of the control flow graph is exposed. These approaches then treat inputs as variables, and branch conditions as constraints on these variables. To find an input that will pass a certain branch (e.g. enter the if condition of an if/else statement), a SAT, or constraint solver is used. By explicitly solving for inputs that satisfy constraints, one can theoretically find an input that will reach any statement in a program. That being said, the downside of such approaches is in their speed, and need for resources. Average programs can have hundreds to thousands of non-trivial branch conditions, and as such, it can take extended periods of time to solve the corresponding equations. This is exacerbated by several factors, including the size of the inputs, reliance on external libraries in the code, and the ease of which one can instrument a given program.

On the flipside, grey-box approaches assume minimal transparency into a program. Rather than reason about the entire control flow graph, for many grey-box approaches, it is enough to instrument the program at run time, solely for the purpose of tracking when an input hits a new (previously unseen) basic block, or a new edge between basic blocks. Approaches like American Fuzzy Lop \citep{AFL}, and it's many variants \cite{Stephens2016DrillerAF, Lemieux2017FairFuzzTR, Bhme2016CoveragebasedGF, Bhme2017DirectedGF} track this information to guide a simple genetic algorithm that generates a series of inputs and executes them through the program.

The key here is speed; generating new inputs to test is near instantaneous, and the only real limiting factor is the speed of execution through the program. In many cases this speed of execution is not negligible - and furthermore, as many of the generated inputs are either redundant, or trivially malformed, AFL and related approaches are wasteful, spending thousands of cycles on inputs that add no new meaningful information. The goal of our work is to introduce a new approach that obtains similar speed and efficiency to AFL, while using machine learning techniques to obtain some of the precision and reasoning ability as white-box approaches. With such an approach, we limit wasted cycles, while retaining the ability to find bugs efficiently and at scale.

\section{Approach}

\begin{algorithm}[t]
  \begin{algorithmic}[1]
      \State \emph{//} \textbf{Algorithm for AFL + Program Modeling}
      \State \emph{// afl: Instance of AFL for generating/executing inputs}
      \State \emph{// iterations: Fixed number of generation iterations}
      \State \emph{// num_generate: New inputs to generate each iteration}
      \State \emph{// $\alpha$: Fraction of generated inputs to execute each iteration}
      \State \emph{// queue: Queue of inputs that exercise new code paths}
      \State \emph{// model: Predicts distribution over execution paths for an input}
      \State \emph{// ranker: Given predictions, ranks by entropy values (high - low)}
      
      \For{$i \in \bf{range(\emph{iterations})}$}
            \State $generated \gets \emph{afl}\tt{.generate(\emph{queue}, \emph{num\_generate})}$
            \State $execute := []$
            \For{$g \in generated$}
                \State $execute \gets \emph{model}\tt{.predict(g)}$
            \EndFor
            \State $execute \gets \emph{ranker}\tt{.rank(\emph{execute})}$
            \For{$j \in \bf{range(\emph{$\alpha \cdot$ num\_generate})}$}
                \State $queue, path \gets \emph{afl}\tt{.execute(\emph{queue, execute[j]})}$
                \State $model \gets model\tt{.retrain(execute[j], path)}$
            \EndFor
      \EndFor
  \end{algorithmic}
  \caption{AFL + Program Modeling Fuzzing Algorithm}
  \label{alg:entropy}
\end{algorithm}

Modeling program behavior is key to improving fuzzing efficiency. While there are many ways to approach this modeling problem, in this work, we focus on learning forward prediction models: given an input, predict the corresponding execution path through the program. If we had a perfect execution model, we could simply skip inputs that lead to execution paths we have already seen, saving significant time.  Our approach, described below, is based on the heuristic that the less confident our model is in the execution path it predicts for a given input, the more likely that input is to lead to an execution path that we have never seen before.

There is an additional benefit to this method. As we execute each input, we get additional training data for the prediction model. Selecting new inputs on the basis of uncertainty is a well-known active learning technique and so this input selection method also serves to hasten prediction model improvement and thus the ability of the system to find good candidates.

At a high level, our approach is to repeatedly perform the following steps:

1) Use AFL to generate some number of possible children inputs, 2) Feed these inputs through our model to predict distributions over execution paths, 3) Rank these generated inputs by the confidence in the predictions, 4) Execute some fraction of those ranked inputs that we are the least confident about, and 5) Use the executed inputs to retrain our path prediction model. This process is graphically depicted in Figure \ref{fig:pipeline}, and logically depicted in Algorithm \ref{alg:entropy}. In the remainder of this section, we provide additional details about each of the steps in the above inner loop: generating candidate inputs, learning a path prediction model, and using this model to rank and execute the candidates.

\subsection{Generating Candidate Inputs}
\label{sec:generation}

To select a set of promising inputs to execute through the program, we first need candidates. We obtain this set of candidates by applying AFL's mutation logic. Specifically, given our input queue, we first sample a parent input, then we apply a set of mutation operators to obtain our new candidate. We repeat this process $K$ times to obtain a full batch of promising candidates. Note that this process is extremely fast, as we do not execute any of the generated inputs through the program.

\subsection{Modeling Programs via Path Prediction}
\label{sec:pathpred}

A crucial component of our approach is to understand program behavior by prediction execution paths from program inputs. When fuzzing starts, we have no examples on which to train our model, so we predict a uniform distribution over a single (null) path for each example, which effectively results in a random ranking over the batch when ranking based on confidence (as will be discussed in Section \ref{sec:execution}). However, after executing the first batch of inputs, from the first iteration of Algorithm \ref{alg:entropy}, we have an initial set of labeled examples $\{x, p\}$, where $x$ corresponds to a featurized representation of the input (e.g. a bag of words representation of an input string), and $p$ represents the corresponding execution path. Note that for our purposes, the execution path $p$ is represented as a unique label (i.e. each observed execution path gets its own label), rather than as a sequence of basic blocks, or edges in the underlying control flow graph. This is because programs can have hundreds of basic blocks, while we note that in practice, only a handful of unique execution paths (sets of traversed basic blocks) are observed.  

With these examples $\{x, p\}$, and a total number of unique observed execution paths $P$, we can then train a probabilistic classifier to predict a distribution over the $P$ paths for a given input $x$. In our experiments, we build the probabilistic classifier using multinomial logistic regression (via the one-vs-all reduction to $P$ separate binary logistic regression models, for efficiency). We featurize our inputs using bigram counts over the bytes of each input string --- that is, we compute a histogram of how many times each unique bigram sequence appears in the input, and use that histogram as our representation. We choose not to utilize an L1 or L2 penalty, and train all models to convergence.

\subsection{Estimating Uncertainty via Entropy}
\label{sec:execution}

The final piece of our approach is using our model to make decisions about which candidates to actually execute through the program. To do this, we apply our hypothesis that inputs with uncertain predictions are more likely to exhibit execution paths that have not been observed, while inputs with confident predictions are more likely to be redundant.

As a measure of uncertainty, we use the entropy of the predicted distribution over execution paths, with high entropy referring to high uncertainty, and low entropy referring to low uncertainty. Given an input $x$, let $\Pr(p_i \mid x)$ be the probability that $x$ exhibits execution path $p_i$. Given this distribution, we compute the entropy as:
\begin{align*}
    H(x) = \sum_{i = 1}^P \Pr(p_i \mid x) \log(\Pr(p_i \mid x))
\end{align*}

With this formula, we then score each generated input in the batch, for the given iteration. Then we rank them by their entropy (highest - lowest). Finally, we select a fraction $\alpha$ of the highest entropy inputs to execute. A full breakdown of the process can be found in Algorithm \ref{alg:entropy}.

\section{Dataset}

We run preliminary experiments on a subset of the DARPA Cyber Grand Challenge Binaries. This dataset consists of 200 separate programs released as part of a 2016 challenge to create tools for finding, verifying, and patching bugs. Many new tools building off of AFL and symbolic execution based approaches came out of this contest \cite{Stephens2016DrillerAF, Cha2015ProgramAdaptiveMF}, and this dataset has been used to benchmark similar tools ever since. Each program provides unique functionality, and was written by humans (affiliated with various DARPA programs). More importantly, each program was written with one or more human-written bugs, meant to mimic errors that developers might make when writing actual programs. Furthermore, the programs in this dataset range in complexity.

In our work, we utilize a subset of 24 randomly chosen programs from this dataset (due to time constraints, we could not run on the full 200). We utilized a version of DARPA CGC binaries compiled for x86 Linux (as opposed to the original DARPA-specific VM), released via this link: \url{https://github.com/trailofbits/cb-multios}.

\section{Experimental Setup}

\begin{figure}[t]
\includegraphics[width=\linewidth]{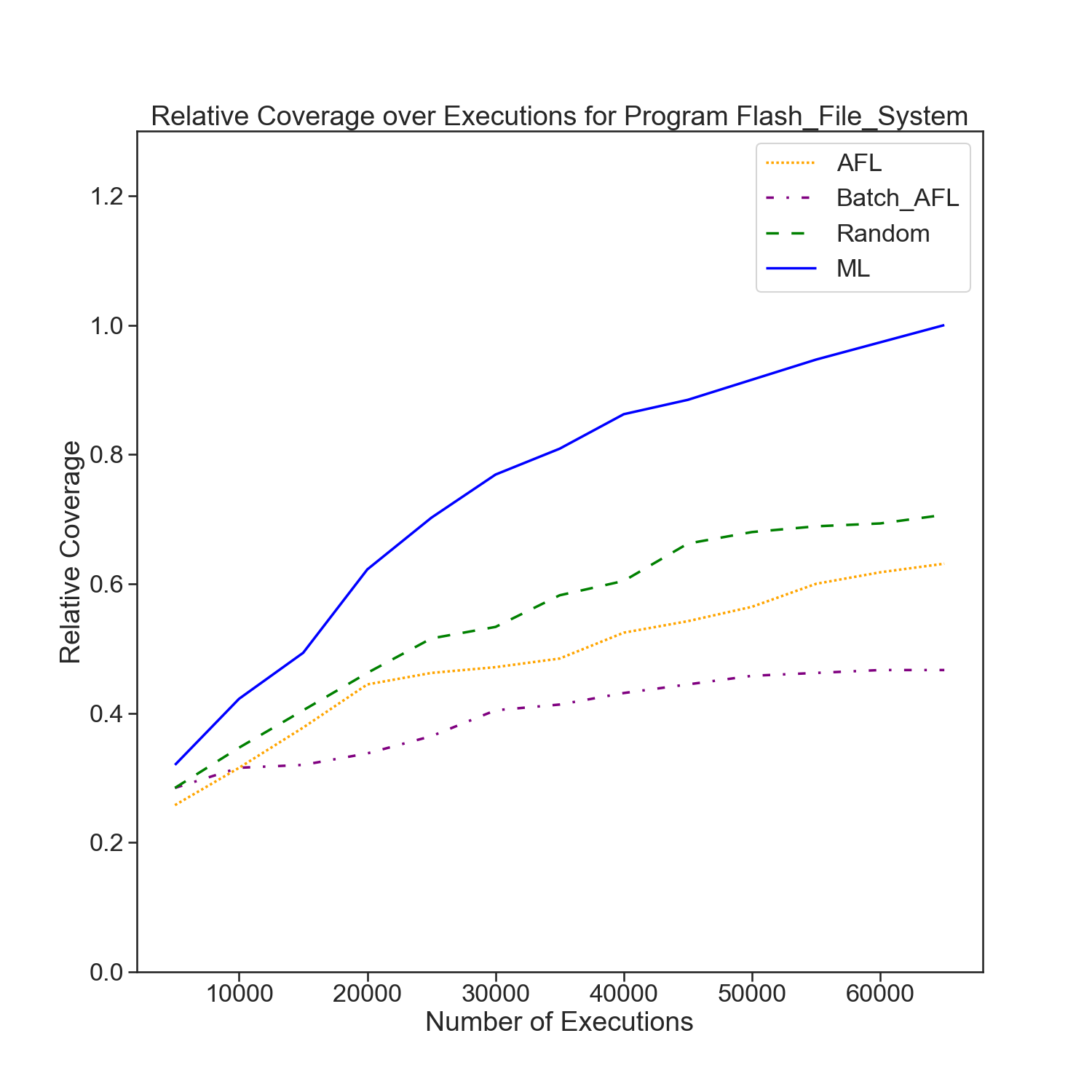}
\caption{Graph depicting relative coverage over number of executions for the program Flash File System (in the CGC binaries). Here, we see that the Program Modeling approach (ML) outperforms all the baselines by a significant margin. Furthermore, as we continue execution, the gap between the ML strategy and the others grows.}
\label{fig:flash}
\end{figure}

We implement our program modeling approach using logistic regression as our prediction model, and featurize our input strings by collecting a bag of byte bigrams (a histogram of the number of times each unique pair of bytes appear in the input). We compare our program modeling approach with three strong baselines. The first baseline is that of AFL itself, as it ships out of the box. However, rather than use the standard AFL parameters, we run AFL with the ``-d'' flag, or ``FidgetyAFL'' \cite{FidgetyAFL, Lemieux2017FairFuzzTR}, as it performs better than standard AFL given a short fuzzing time period. 

The second baseline we utilize is a batched version of AFL, which we refer to as Batched FidgetyAFL. The differences between Batched FidgetyAFL and FidgetyAFL are as follows: FidgetyAFL updates its state (its queue, and therefore which parent inputs are sampled to generate the next child) immediately after each input is generated and executed. The batched versions instead remove this consistent state update and replace it with a batched update, where multiple inputs are first generated all together without any state updates, and then executed (with state updates) all at once. We choose this baseline as it offers a better comparison to our program modeling approach. Recall that in our approach, we first generate a batch of examples (without execution), then rank the inputs among the batch to pick the fraction $\alpha$ to execute. Like Batched FidgetyAFL, we do not update the state of AFL's queue until after we execute all the ranked inputs.

In addition to FidgetyAFL and Batched FidgetyAFL, we have a third baseline, Random Batched FidgetyAFL, which explores the effect the ranking over the generated inputs has relative to fuzzing performance. Unlike our program modeling approach, which generates a large number of inputs, then executes the top $\alpha$-fraction after ranking the predictions by their entropy, Random Batched FidgetyAFL randomly picks a fraction $\alpha$ of the batch to execute. In this way, this baseline lets us examine if our entropy-ranking approach is actually working. Note that Random Batched FidgetyAFL is markedly different from the Batched FidgetyAFL baseline --- this is because AFL does not uniformly sample inputs from its queue --- instead, it uses a heuristic \textit{schedule} to sample queue inputs, first sampling elements from the queue that are more recent, then later (after generating many new inputs) starts sampling other elements from the queue. In this way, Random Batched FidgetyAFL exhibits slightly more random behavior than Batched FidgetyAFL, as it reflects a wide variety of different sampled parents. 

\begin{table*}[t]
\begin{tabular}{@{}lcccc@{}}
\toprule
Executions & FidgetyAFL                          & Batched FidgetyAFL                   & Random Batched FidgetyAFL                      & Logistic Regression with Bigram Features        \\ \midrule
10000      & .623 $\pm$ .011 & .624 $\pm$ .011 & .632 $\pm$ .011 & \textbf{.638 $\pm$ .011} \\
20000      & .647 $\pm$ .011 & .644 $\pm$ .011 & .667 $\pm$ .011 & \textbf{.688 $\pm$ .011} \\
30000      & .671 $\pm$ .011 & .660 $\pm$ .011 & .699 $\pm$ .011 & \textbf{.755 $\pm$ .009} \\
40000      & .692 $\pm$ .010 & .671 $\pm$ .010 & .716 $\pm$ .011 & \textbf{.791 $\pm$ .009 }\\
50000      & .706 $\pm$ .010 & .680 $\pm$ .010 & .755 $\pm$ .009 & \textbf{.816 $\pm$ .009} \\ \bottomrule
\end{tabular}
\caption{Aggregate statistics over 24 programs. FidgetyAFL is the out-of-the-box release of FidgetyAFL (standard AFL run in ``havoc'' mode), Batched FidgetyAFL operates on batches of inputs, choosing the first $\alpha$ fraction to execute, and Random Batched FidgetyAFL also operates on batches, but randomly chooses which $\alpha$ fraction to execute. Logistic Regression with Bigram Features implements the procedure in Algorithm \ref{alg:entropy}, and uses a learned model of program behavior to rank inputs to execute.} 
\label{tbl:results}
\end{table*}

We run our experiments on 24 of the DARPA CGC binaries, for a total of 50,000 executions per binary. To jumpstart learning, and to eliminate most of the variance across fuzzing runs, we start all runs by letting FidgetyAFL run for a 3 minute period. We pre-train our logistic regression model on the inputs executed during this window. We then use the resulting AFL state, and the queue of inputs created as our initial queue, and start each of the 4 different strategies on top (FidgetyAFL, Batched FidgetyAFL, Random, and Logistic Regression). For all experiments, we utilize AFL version 2.52b.

\begin{figure}[b]
\includegraphics[width=\linewidth]{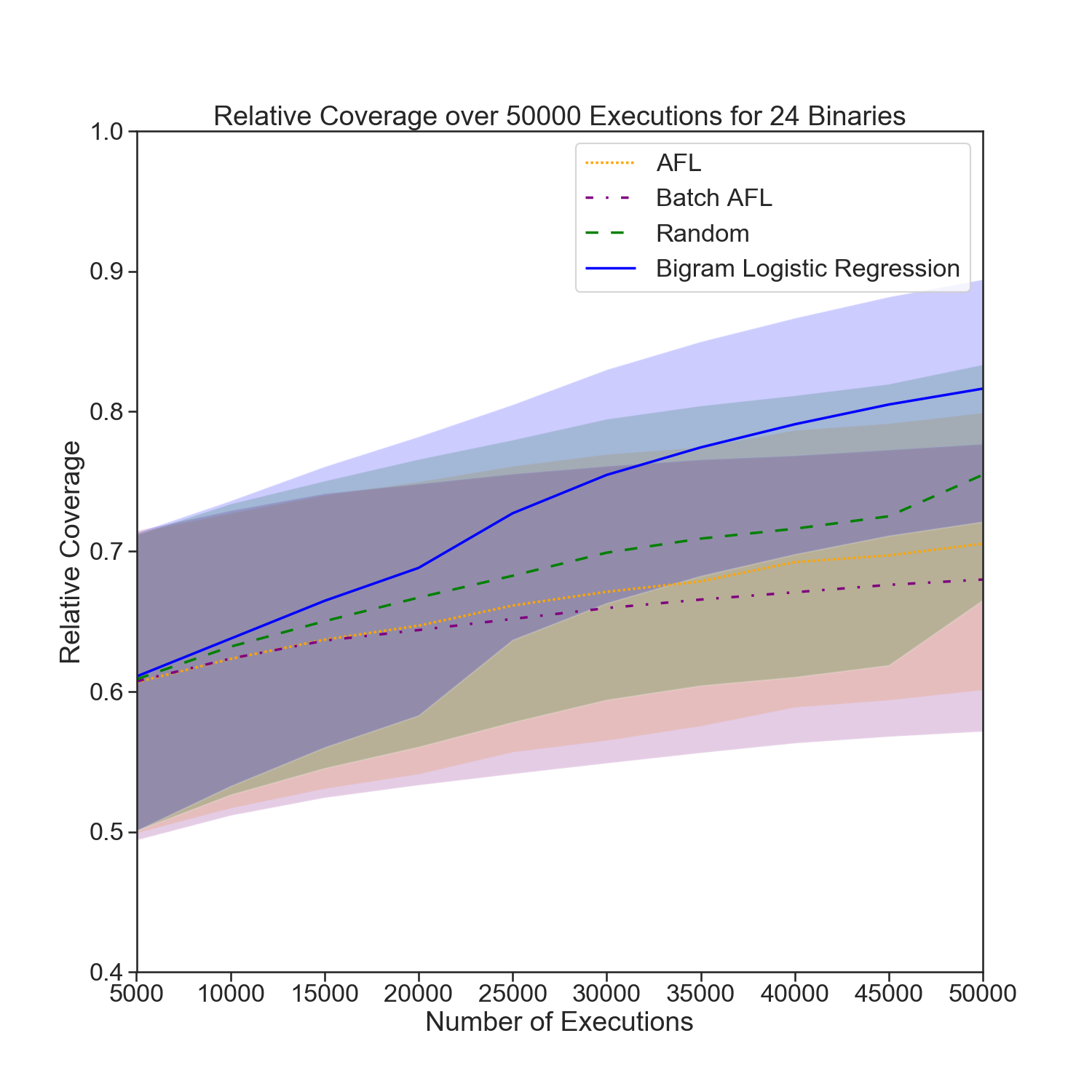}
\caption{Summary Graph depicting relative coverage over number of executions for the 24 CGC binaries, at a 95\% confidence interval. Again, we see the Program Modeling approach (ML) outperform the other approaches by a larger and larger margin as execution continues.}
\label{fig:summary}
\end{figure}

To measure the relative performance across all strategies, we use a metric we refer to as relative coverage. Let $s$ correspond to a given strategy, $t$ the given execution iteration, $T$ the max number of execution iterations, and \textit{$\text{code-paths}_t(s)$} the number of unique code-paths that strategy $s$ has discovered by execution iteration $t$. Then Relative Coverage \textit{$\text{rel-cov}$} for a single program is defined as:
\begin{align*}
	\text{rel-cov}_t(s) &= \dfrac{\text{code-paths}_t(s)}{\max_{s'} [\text{code-paths}_T(s')]}
\end{align*}
or the ratio between the number of code paths strategy $s$ has found, and the maximum number of code paths across all strategies by the final execution iteration $T$. We report the mean and standard error of across all 24 programs. Furthermore, to get a better sense of how the different strategies behave over time, we report relative coverage statistics at every 10,000 executions.

\section{Results and Discussion}

Table \ref{tbl:results} reports relative coverage statistics for each of the four strategies at each interval of 10,000 executions. Furthermore, Figure \ref{fig:flash} provides a graph reporting number of code paths discovered vs. executions for program Flash File System (an example program from the CGC binaries). Finally, Figure \ref{fig:summary} contains a summary graph aggregating relative coverage over all 24 binaries in our test set, at a 95\% confidence interval across binaries. 

From these results, there are two key conclusions to pick out. The first is to realize that at all time steps, the program modeling approach is a clear winner, obtaining higher coverage than any of the baseline strategies. This seems to indicate that the gains from Logistic Regression ranking are significantly higher than the losses suffered from the batched update procedure. As such, a possible avenue for future work would be to augment the Entropy Ranking based Logistic Regression with a thresholding operation, to allow the model to make choices about whether to execute an input or not, in an online fashion. Doing so would remove any need for batching, and allow the approach to incur the same benefits as traditional AFL, with its continuous state updates.

The second key observation is that the performance gap between the program modeling approach and the other baseline approaches grows as the number of executions rises. This is best exhibited by the graphs in Figures \ref{fig:flash} and \ref{fig:summary}. We see that at the beginning of fuzzing, there is just a small, almost negligible gap in performance, while as execution continues, the gap grows larger and larger. There are two possible conclusions to be drawn from this: the first is a rather simple one, that as the program modeling approach identifies more code paths, AFL's queue is updated, and we begin sampling more of the recently discovered inputs --- reasons for why AFL itself is successful. However, another possible explanation is that of how the program model behaves as more data becomes available. With more executions, the logistic regression is given more labeled examples. As such, it gets better at identifying patterns in the inputs, and the confidence scores assigned at inference time become more meaningful. Another avenue of future work is to examine the nature of the learned models, and how they change as execution continues. It may also be worthwhile to throw stronger learning algorithms and more program-specific features into the mix, seeing if there is a way to strengthen the reported confidence scores.

\section{Conclusion}
 
In this work, we presented a system for improving the efficiency and precision of AFL, the premier grey-box fuzzer, utilizing techniques from machine learning to directly model program behavior. Specifically, we note that a major weakness of AFL and similar approaches is the number of program executions that are wasted on redundant or non-informative inputs. To remedy this problem, we proposed a two-phase approach that 1) learns a forward prediction model that maps inputs to execution paths, and 2) uses that model to identify inputs that are potentially interesting. The intuition we use is that if we can confidently model how a given input will behave when executed, then it is not worth executing. Instead, we should focus on the inputs for which our model exhibits low confidence --- these are inputs that when executed, will potentially trigger new areas of code that have not yet been observed. Our results show that our ranking-based approach built with a simple logistic regression classifier obtains extremely strong performance, beating 3 strong baselines, including the standard, out-of-the-box implementation of AFL itself. Furthermore, our results show that as we continue fuzzing, our approach gets better and better, with the performance gap between our approach and baselines widening over time. These results indicate that there are strong benefits to be found in applying techniques from machine learning and pattern recognition to fuzzing, and that this is a very fruitful avenue of research.

\bibliographystyle{ACM-Reference-Format}
\bibliography{references}

\end{document}